\documentclass[conference]{IEEEtran}
\IEEEoverridecommandlockouts
\usepackage{cite}
\usepackage{amsmath,amssymb,amsfonts}
\usepackage{algorithmic}
\usepackage{graphicx}
\usepackage{textcomp}
\usepackage{xcolor}
\usepackage{booktabs}
\usepackage{CJKutf8}

\def\BibTeX{{\rm B\kern-.05em{\sc i\kern-.025em b}\kern-.08em
		T\kern-.1667em\lower.7ex\hbox{E}\kern-.125emX}}

\begin{document}
\begin{CJK*}{UTF8}{gbsn}
	
	\title{Curriculum Recommendations Using Transformer Base Model with InfoNCE Loss and Language Switching Method\\ 
	}
	
        \author{\IEEEauthorblockN{ Xiaonan Xu}
        \IEEEauthorblockA{
        \textit{Northern Arizona University}\\
        Flagstaff, USA \\
        xiaonan.xu.academic@outlook.com}
        \and
        \IEEEauthorblockN{ Bin Yuan}
        \IEEEauthorblockA{
        \textit{Trine University}\\
        Phoenix, USA \\
        binyuan2235@gmail.com}
        \and
        \IEEEauthorblockN{ Tianbo Song\textsuperscript{*}}
        \IEEEauthorblockA{
        \textit{Arizona State University}\\
        Phoenix, USA \\
        tianbosong@aol.com}
        \and
        \IEEEauthorblockN{ Shulin Li}
        \IEEEauthorblockA{
        \textit{Trine University}\\
        Phoenix, USA \\
        liam.cool666@gmail.com}
        }

	\maketitle
	
	\begin{abstract}

        The Curriculum Recommendations paradigm is dedicated to fostering learning equality within the ever-evolving realms of educational technology and curriculum development. In acknowledging the inherent obstacles posed by existing methodologies, such as content conflicts and disruptions from language translation, this paradigm aims to confront and overcome these challenges. Notably, it addresses content conflicts and disruptions introduced by language translation, hindrances that can impede the creation of an all-encompassing and personalized learning experience. The paradigm's objective is to cultivate an educational environment that not only embraces diversity but also customizes learning experiences to suit the distinct needs of each learner. By proactively identifying and addressing these issues, the paradigm strives to pave the way for a more inclusive and responsive educational landscape, ensuring that learning opportunities are equitable and tailored to individual learners' requirements.
        
        To overcome these challenges, our approach builds upon notable contributions in curriculum development and personalized learning, introducing three key innovations. These include the integration of Transformer Base Model to enhance computational efficiency, the implementation of InfoNCE Loss for accurate content-topic matching, and the adoption of a language switching strategy to alleviate translation-related ambiguities. Together, these innovations aim to collectively tackle inherent challenges and contribute to forging a more equitable and effective learning journey for a diverse range of learners. Competitive cross-validation scores underscore the efficacy of sentence-transformers/LaBSE, achieving 0.66314, showcasing our methodology's effectiveness in diverse linguistic nuances for content alignment prediction.

	\end{abstract}
	
	
        \section{Introduction}

        In the ever-changing landscape of educational technology and curriculum development, the imperative pursuit of learning equality takes center stage. The commitment to delivering an inclusive and effective learning experience across diverse domains has spurred innovative approaches in curriculum recommendations. This article delves into the paradigm of Learning Equality - Curriculum Recommendations, recognizing the profound impact of curriculum recommendations on shaping the learning journey. At its core, the paradigm strives to cultivate an educational environment that caters to the unique needs of learners, ensuring equal opportunities for all. The overarching goal is to render the learning experience not only personalized but also equitable, acknowledging and surmounting the challenges posed by existing methodologies. Through the introduction of novel strategies, the paradigm addresses these challenges head-on, aiming to revolutionize the educational landscape by fostering an environment where every individual's educational journey is both personalized and marked by a commitment to equality.

        In the realm of educational technology and curriculum development, a multitude of significant contributions have shaped the learning landscape. Tucker's K-12 computer science model\cite{tucker2003model} and Chamunyonga et al.\cite{chamunyonga2020impact} call for enhanced medical curricula underscore impactful contributions in educational technology. Moreover, Bian et al.\cite{bian2021contrastive} addressed predictive model robustness, Wang et al.\cite{wang2022research} tackled personalized learning nuances, Kumar et al.\cite{kumar2022customized} navigated customized approaches, and Marras et al.\cite{marras2022equality} confronted bias and fairness challenges. In the latest research conducted in 2023, Sanusi et al.\cite{sanusi2023systematic} provided a comprehensive overview of machine learning in K-12 education, Hassan et al.\cite{hassan2023leveraging} used deep learning to enhance computing curriculum, and Atalla et al.\cite{atalla2023intelligent} introduced an intelligent academic advising system. However, despite these impactful contributions, challenges such as potential biases, fairness issues, and the need for more tailored strategies persist in educational technology and curriculum development.
        
        Despite these commendable efforts, there are inherent issues within existing methodologies. Challenges include handling related content conflicts, addressing noise in training data introduced by language translation, and ensuring effective sampling strategies to mitigate false positives in content-topic associations. Our approach introduces three key innovations to tackle the identified challenges effectively. First, we employ Transformer Base Model\cite{vaswani2017attention}\cite{feng2020language} with  the InfoNCE Loss\cite{wu2021rethinking} as a symmetric contrastive loss-function. This facilitates precise loss calculation by emphasizing correct matches on the diagonal of the similarity matrix, reducing noise in the training process. Secondly, to mitigate issues introduced by language translation, we implement a language switching strategy after each epoch, involving alternating between languages. This strategy minimizes ambiguities during loss calculation and provides a score boost by diversifying the training data distribution.

        In conclusion, the Learning Equality - Curriculum Recommendations paradigm builds upon the foundations laid by prior works while innovatively addressing their inherent challenges. By implementing Transformer Base Model with limited sequence length, utilizing InfoNCE Loss, and incorporating a language switching strategy, our approach aims to create a more equitable and effective learning environment for diverse learners. Competitive cross-validation scores underscore the efficacy of sentence-transformers/LaBSE, achieving 0.66314, showcasing our methodology's effectiveness in diverse linguistic nuances for content alignment prediction.

	\section{RELATED WORK}

        In the realm of educational technology and curriculum development, several significant contributions have been made to enhance the learning experience across various domains. Tucker \cite{tucker2003model} proposed a model curriculum for K-12 computer science, providing a foundational framework for computer science education at the primary and secondary levels. Meanwhile, Chamunyonga et al.\cite{chamunyonga2020impact} explored the impact of artificial intelligence and machine learning in radiation therapy, emphasizing the need for curriculum enhancement to address evolving technological landscapes in the medical field.

        Due to the rapid advancements in deep learning, an in-
        creasing number of studies are being employed in the real of Curriculum Recommendations research. Bian et al.\cite{bian2021contrastive} introduced the concept of contrastive curriculum learning for sequential user behavior modeling, employing data augmentation techniques to improve the robustness of predictive models. Wang et al.\cite{wang2022research} delved into online learner modeling and course recommendation based on emotional factors, adding a nuanced dimension to personalized learning approaches. Kumar et al.\cite{kumar2022customized} focused on customized curriculum and learning approach recommendation techniques, specifically in the application of virtual reality in medical education, highlighting the importance of tailoring educational strategies to emerging technologies. Marras et al.\cite{marras2022equality} investigated the equality of learning opportunities through individual fairness in personalized recommendations, addressing the challenges associated with bias and fairness in educational technology. 
        
        In the latest research conducted in 2023, the exploration of the Learning Equality - Curriculum Recommendations has continued to evolve. Sanusi et al.\cite{sanusi2023systematic} conducted a systematic review of teaching and learning machine learning in K-12 education, providing a comprehensive overview of the current landscape and identifying potential areas for improvement. Hassan et al.\cite{hassan2023leveraging} leveraged deep learning and big data to enhance computing curriculum for industry-relevant skills, presenting a Norwegian case study that underscores the integration of cutting-edge technologies into educational practices. Finally, Atalla et al.\cite{atalla2023intelligent} introduced an intelligent recommendation system for automating academic advising based on curriculum analysis and performance modeling, showcasing innovative approaches to support academic decision-making.

	\section{ALGORITHM AND MODEL}
  
        \subsection{Transformer Base Model with InfoNCE Loss} 
        In this section, we will delve into our points of innovation. By implementing a Transformer Base Model with a limited sequence length, utilizing InfoNCE Loss, and incorporating a language switching strategy, our approach aims to create a more equitable and effective learning environment for diverse learners.
        
\begin{figure}[htbp]
        \centering
        \includegraphics[width=1\linewidth]{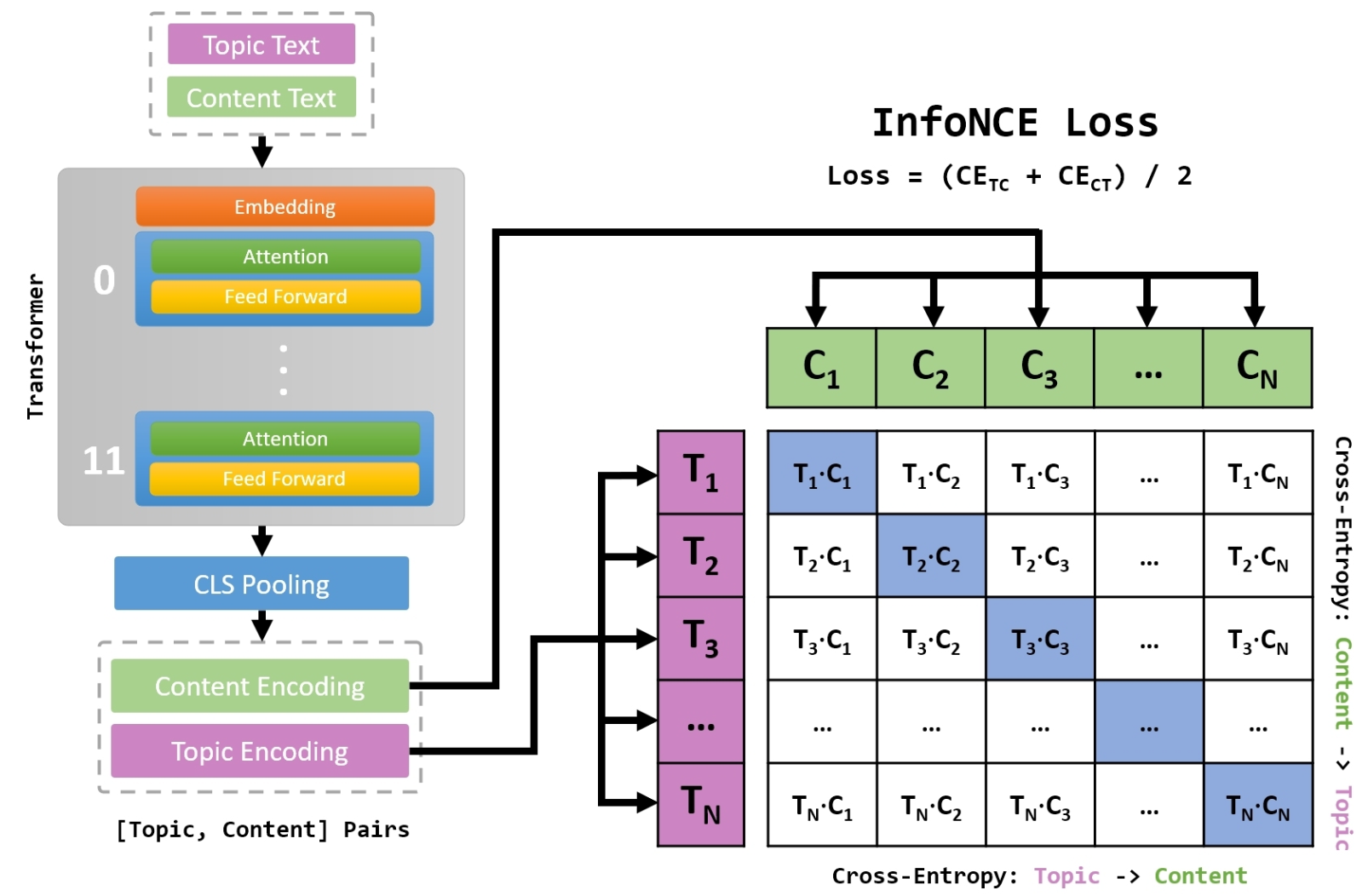}
        \caption{InfoNCE Loss}
        \label{fig:1}
        \end{figure}

        To address challenges in topic-content matching, we employ a Transformer Base Model with the InfoNCE Loss, as detailed in Vaswani et al.\cite{vaswani2017attention} and Feng et al.\cite{feng2020language}. Illustrated in Fig.\ref{fig:1}, this architecture enables nuanced learning of topic-content relationships. Leveraging Transformer Base Models to encode both topics and content, the design empowers the model to capture intricate patterns and dependencies within the input data.
        
        By utilizing the InfoNCE Loss\cite{wu2021rethinking}, our model aims to reduce noise in the training process, particularly when dealing with topics that share similar content. Unlike traditional cross-entropy loss, InfoNCE Loss emphasizes correct matches on the diagonal of the similarity matrix during loss calculation. This innovative loss function contributes to the model's ability to precisely identify and match topics with their corresponding content. The InfoNCE Loss is formulated as follows:
        
        \begin{equation}
        -\log\left(\frac{\exp(\text{sim}(x, y^+))}{\exp(\text{sim}(x, y^+)) + \sum_{k=1}^{K} \exp(\text{sim}(x, y_k^-))}\right) 
        \end{equation}
        Where:
        \begin{itemize}
            \item \(\text{sim}(x, y)\) is a similarity measure between samples \(x\) and \(y\).
            \item \(y^+\) is the positive sample similar to \(x\).
            \item \(y_k^-\) is the \(k\)-th negative sample dissimilar to \(x\).
            \item \(K\) is the number of negative samples.
        \end{itemize}
        
        This loss function guarantees that, during the loss calculation, emphasis is placed solely on the diagonal of the similarity matrix. This is vital to prevent misinterpretation of elevated similarities between topics and content within the same batch. By focusing exclusively on the diagonal, the loss function ensures a more accurate assessment of matching within the context of the specific batch, mitigating the risk of erroneous interpretations of similarities between disparate elements in the dataset.

        Moreover, we integrate a specialized shuffle function to construct batches that prevent the co-occurrence of topics and related content. This approach is essential for minimizing noise and conflicts in the training process. Additionally, we employ a sampling strategy for missing and incorrect content, ensuring that the model effectively learns from both positive and challenging negative samples. This comprehensive approach enhances the model's robustness by systematically addressing potential biases and difficulties inherent in the training data, promoting a more accurate and well-rounded learning experience.

        \subsection{Language Switching Method}
        We adopted a translation approach covering prevalent languages called language switching method. However, directly incorporating translated data into our training set presents challenges with the InfoNCE loss. For instance, translating an English topic and content item into French, where a corresponding topic and content already exist in the original French data, would introduce noise in loss calculation. To address this, we implement a careful filtering mechanism to exclude translated pairs with similar counterparts in the target language, ensuring that the augmented data contributes meaningfully to the training process without introducing redundancies or misleading information during loss calculation.

\begin{figure}[htbp]
        \centering
        \includegraphics[width=1\linewidth]{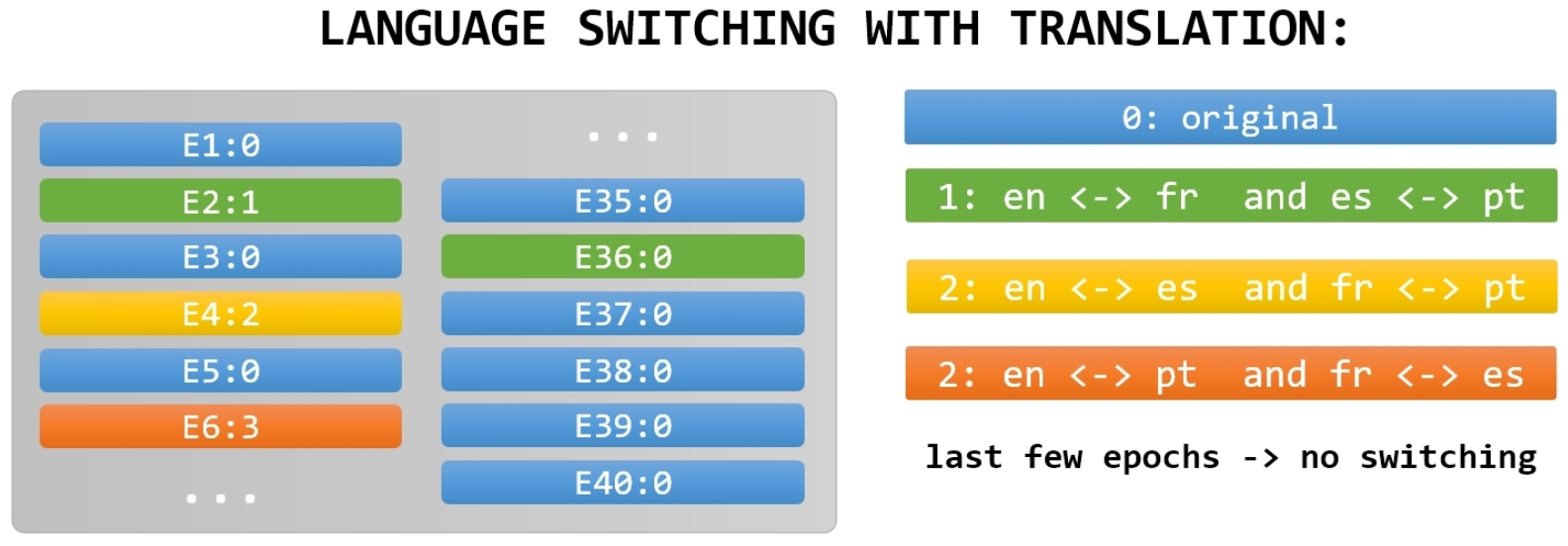}
        \caption{Language Switching Method}
        \label{fig:2}
        \end{figure}

        As shown in Fig.\ref{fig:2}, the switching strategy occurs every second epoch and is confined to languages like English (en), Spanish (es), Portuguese (pt), and French (fr). Despite anticipated noise in training, language switching contributes to a score increase of approximately 0.01–0.02. This modest gain implies that the introduced noise might influence training dynamics positively. The periodicity of the switching strategy, coupled with its language limitations, strategically introduces variability during training, and the observed score boost suggests a nuanced impact on the model's ability to adapt and generalize, showcasing the strategy's subtle yet meaningful role in enhancing the overall training process.
        
            \subsection{Datasets}
        
        The dataset originates from the Kolibri Studio curricular alignment tool, allowing users to create channels, construct topic trees representing curriculum taxonomies, and organize content items. Users can upload their content or import materials from the Kolibri Content Library. The challenge is to predict the optimal alignment of content items with topics, mimicking the choices made by curricular experts and Kolibri Studio users. The widespread integration of AI is evident\cite{mou2023research, dai2023integrative, mou2023automated}. The goal is to streamline the curator's task of discovering pertinent materials for each topic by recommending content items. The comprehensive test set comprises 10,000 additional topics (absent in the training set) and numerous extra content items exclusively correlated with the test set topics. This expansion aims to assess the model's ability to generalize and recommend relevant content beyond the training data, simulating real-world scenarios where novel topics and content continuously emerge.
                
            \subsection{Evalution metrics}

        The $F_{2}$ score\cite{prasetiyo2021evaluation}, also known as the F-beta score, serves as a valuable metric for assessing the accuracy of a test. It represents the weighted harmonic mean of precision and recall, with an emphasis on recall. The general formula for the F-beta score is expressed as:

        \begin{equation}
        F_{\beta} = \frac{(1 + \beta^2) \cdot \text{precision} \cdot \text{recall}}{\beta^2 \cdot \text{precision} + \text{recall}}
        \end{equation}
        
        Here, precision is defined as \(\frac{TP}{TP + FP}\), and recall is defined as \(\frac{TP}{TP + FN}\), where TP is the number of true positives, FP is the number of false positives, and FN is the number of false negatives.
        
        For the $F_{2}$ score, \(\beta = 2\), indicating that recall is considered twice as important as precision. Substituting these values into the F-beta formula, we derive the following formula for the $F_{2}$ score:
        
        \begin{equation}
        F_{2} = \frac{5 \cdot TP^2}{TP^2 + 5 \cdot TP \cdot FP + TP \cdot FN}
        \end{equation}
        
        To obtain the mean $F_{2}$ score across all predictions in a dataset, calculate the $F_{2}$ score for each row and then compute the average. Additionally, a 10-fold cross-validation (CV)\cite{browne2000cross} split strategy is employed, aiming to minimize the overlap of content relations between different folds. This involves creating 10 buckets and assigning topics to the bucket where all attached content results in the least overlap with the same attached content in other buckets. While achieving a perfect split aligned with the leader board is challenging due to the n x m relation of Topic x Content, efforts have been made to minimize the overlap during CV-split creation.
            
            \subsection{Results}
        In this section, we evaluate the performance of our model using a diverse set of pre-trained models to predict content alignments for given topics. Notable models such as sentence-transformers/LaBSE\cite{feng2020language}, facebook/mcontriever-msmarco\cite{izacard2021towards}, sentence-transformers/stsb-xlm-r-multilingual, and sentence-transformers/paraphrase-multilingual-mpnet-base-v2\cite{reimers2019sentence} were employed. Training involved fold 0 of 10 Folds, with a max sequence length of 96 and a batch size of 768 pairs of [topic, content]. Utilizing polynomial decay with warmup (2 epochs) for 40 epochs and a learning rate of 0.0003. Table \ref{tab:model-results} presents the cross-validation scores, showcasing competitive performance across the models.
        \begin{table}[h]
            \centering
            \caption{Model Results}
            \label{tab:model-results}
            \begin{tabular}{ccc}
                \toprule 
                \textbf{Model\_Name} & \textbf{CV Score} \\
                \midrule 
                sentence-transformers/LaBSE & 0.66314 \\
                facebook/mcontriever-msmarco & 0.66148 \\
                sentence-transformers/stsb-xlm-r-multilingual & 0.65697 \\
                sentence-transformers/paraphrase-multilingual-mpnet-base-v2 & 0.66035 \\
                \bottomrule 
            \end{tabular}
        \end{table}
        
        The competitive cross-validation scores highlight the efficacy of our sentence-transformers/LaBSE employing the InfoNCE loss and Language Switching leading with a score of 0.66314. These outcomes underscore the effectiveness of our methodology in capturing diverse linguistic nuances for content alignment prediction. The ensemble approach, incorporating innovative techniques, proves to be a robust solution for the task, showcasing its ability to yield competitive results in comprehensively addressing the challenges associated with content alignment prediction in the context of diverse linguistic nuances.
        
        \section{Conclusion}

        Amidst this swift advancement, our study explores the nuanced domain of the Learning Equality - Curriculum Recommendations paradigm, representing a transformative shift in educational technology committed to ensuring inclusive and effective learning experiences. Our innovative approach, integrating Transformer Base Models, the InfoNCE Loss, and a language switching strategy, directly addresses challenges in the paradigm. The competitive cross-validation scores highlight the efficacy of our sentence-transformers/LaBSE employing the InfoNCE loss and Language Switching leading with a score of 0.66314. These outcomes underscore the effectiveness of our methodology in capturing diverse linguistic nuances for content alignment prediction. 
        
        Our study contributes to the ongoing evolution of AI's role \cite{xiao2022dual, wang2020mrmrp, zeng2022graph} in fostering a progressive and inclusive educational landscape. Amalgamating for educational advancement, we ensure tech fosters fairness and efficacy in dynamic learning frontiers. Navigating curriculum complexities, Learning Equality's paradigm is a beacon, championing personalized and equitable learning experiences in the educational landscape.

        \bibliographystyle{IEEEtran}
        \bibliography{references}
\end{CJK*}	
\end{document}